# Comparative Evaluation of Deep Learning Models for Fake Image Detection

Akhitha Pakala[1], Mohammed Mahir Rahman[1][0009-0007-4167-1475], Shahzad Memon[1][0000-0003-3354-5798] and Tauseef Ahmed[1][0000-0002-1850-3496]

[1] University of East London, London E16 2RD, UK
T.Ahmed4@uel.ac.uk

**Abstract -** The growing sophistication of GAN-based image manipulation presents significant challenges for digital forensics. This study compares the performance of four pre-trained CNN architectures—VGG16, ResNet50, EfficientNetB0, and XceptionNet—for fake image detection using a unified preprocessing and training pipeline. A dataset of real and manipulated images was processed through resizing, normalization, and augmentation to address class imbalance and improve generalization. Models were evaluated using Accuracy, Precision, Recall, F1-score, and ROC-AUC. VGG16 achieved the highest accuracy at 91%, with XceptionNet, ResNet50, and EfficientNetB0 each reaching 90%. EfficientNetB0 showed stronger sensitivity to fake images but reduced reliability on real samples, reflecting imbalance-driven bias. Limitations include dataset imbalance, overfitting, and limited interpretability, which affect cross-domain robustness. The study provides a reproducible baseline and underscores the need for balanced datasets, advanced augmentation, and fairness-aware training to develop reliable fake image detection systems.

## 1 INTRODUCTION

The widespread use of digital images in communication, journalism, and social media has made visual content a cornerstone of information exchange. However, the increasing sophistication of image manipulation techniques, particularly those enabled by generative adversarial networks (GANs), has raised serious concerns regarding authenticity, privacy, and societal trust [1]. Manipulated images are frequently used to spread misinformation, damage reputations, commit fraud, and influence public opinion, posing risks to democratic processes and national security [1]. Traditional detection methods, such as watermark analysis, metadata inspection, and pixel-level error detection, are often ineffective against high-quality manipulations produced by modern AI-based synthesis tools [2]. Deep learning, particularly convolutional neural networks (CNNs), has emerged as a promising solution due to its ability to automatically extract features and generalize across complex patterns. Pre-trained architectures such as ResNet50, EfficientNet, and XceptionNet have demonstrated strong performance in diverse computer vision tasks, including object recognition, medical imaging, and facial analysis [3]. Their application to fake image detection enables the identification of subtle artifacts that are imperceptible to the human eye or overlooked by traditional forensic methods. Despite these advances, challenges remain. Deep learning models are prone to overfitting, limited generalization across datasets, and susceptibility to adversarial attacks [4]. Model performance is further influenced by data quality and training strategies. Preprocessing techniques such as resizing, normalization, and augmentation, combined with optimization strategies like dropout, early stopping, and learning rate scheduling, have been shown to improve robustness and accuracy [5]. This study implements and evaluates multiple CNN architectures: VGG16, ResNet50, EfficientNetB0, and XceptionNet; for fake image detection. By comparing their performance under standardized preprocessing and optimization conditions, the research aims to identify trade-offs in accuracy, efficiency, and reliability. Beyond technical evaluation, the study also emphasizes ethical considerations, including fairness, transparency, and interpretability, which are critical for deploying trustworthy detection systems in real-world scenarios [6]. The research is guided by the following objectives:

a. to implement and evaluate pre-trained CNN models for fake image classification,
b. to enhance performance through preprocessing and optimization strategies, and
c. to analyse existing detection methods to highlight challenges and research gaps.

By addressing these aims, the study contributes to the development of reproducible, ethical, and deployable frameworks for fake image detection.

## 2 Related Work

The detection of fake and manipulated images has become a critical research area in digital forensics and computer vision. Early approaches relied on traditional techniques such as watermark analysis, metadata inspection, and pixel-level error detection, but these methods often fail against high-quality manipulations produced by modern generative adversarial networks (GANs) [1], [2]. As a result, deep learning models, particularly convolutional neural networks (CNNs), have emerged as the dominant paradigm for fake image detection. CNNs have demonstrated strong performance in image classification tasks due to their ability to capture low-level features such as edges and textures. Liu and Deng [7] highlighted the role of CNNs in differentiating real and manipulated images, while Rössler et al. [8] showed that subtle pixel-level anomalies introduced by GANs can be effectively detected using CNN-based models. However, shallow CNNs often suffer from overfitting and limited generalization to unseen manipulations [9]. ResNet50 introduced residual connections that enable deeper networks to be trained without degradation, improving robustness against contextual inconsistencies [10]. Studies such as Dang et al. [11] demonstrated that fine-tuned ResNet50 models trained on Celeb-DF datasets achieved accuracy exceeding 98%, outperforming conventional machine learning methods. EfficientNet, proposed by Tan and Le [12], employs compound scaling strategies to balance depth, width, and resolution, offering efficient trade-offs between accuracy and computational cost. Rana and Sung [13] demonstrated that ensembles combining ResNet50 and EfficientNet achieved accuracy above 99% on FaceForensics++ datasets. EfficientNet has also been shown to perform well on resource-constrained devices, making it suitable for real-time deployment [14]. However, domain shifts and subtle semantic inconsistencies remain challenges for its generalization [15]. XceptionNet, introduced by Chollet [16], leverages depth-wise separable convolutions to improve computational efficiency while maintaining expressiveness. Rössler et al. [8] reported that XceptionNet achieved an AUC greater than 0.95 on the FaceForensics++ benchmark, outperforming other CNN architectures. Hybrid approaches have also emerged, such as Ganguly et al. [17], who integrated transformer blocks with XceptionNet to detect multimodal deepfakes involving synchronized audio-visual manipulations. Ritter et al. [18] further compared XceptionNet, ResNet, and EfficientNet, finding XceptionNet superior in detecting frame-level inconsistencies. Despite these advances, several challenges persist. CNN-based detectors often struggle with cross-dataset generalization, adversarial robustness, and explainability [4], [19]. Wang et al. [20] suggested that hybrid architectures combining CNNs with transformers or attention mechanisms could improve contextual understanding. Babu et al. [21] emphasized the need for models capable of adapting to emerging forgery techniques such as latent diffusion and neural radiance fields (NeRFs). Verdoliva [19] highlighted the importance of explainable AI (XAI) tools to ensure transparency and trust in forensic applications. Although ResNet50, EfficientNet, and XceptionNet have shown strong performance in deepfake detection, prior research often evaluates them on different datasets, preprocessing schemes, and training setups, limiting comparability. Moreover, robustness, fairness, and interpretability are seldom assessed systematically across architectures. This work fills that gap by benchmarking four CNN models within a unified experimental framework to provide a reproducible and domain-relevant baseline.

## 3 Methodology

The study adopts an experimental research design to evaluate the performance of multiple pre-trained convolutional neural networks (CNNs) in detecting fake images. The models selected: VGG16, ResNet50, EfficientNetB0, and XceptionNet - were chosen due to their proven success in image classification tasks and prior application in digital forensics [3], [8], [12], [16]. The design emphasizes reproducibility, with standardized preprocessing and optimization strategies applied across all models to ensure fair comparison. The dataset comprised real and manipulated images, including samples generated through advanced editing tools and GAN-based synthesis methods [1], [8]. Images were pre-processed through resizing, normalization, and augmentation to improve generalization and mitigate dataset imbalance [5], [13]. Augmentation techniques included rotation, flipping, brightness adjustment, and noise injection, reflecting real-world variability in manipulated content [22]. Model training and evaluation were conducted using supervised learning. Each CNN architecture was fine-tuned on the dataset, with performance assessed using Accuracy, Precision, Recall, F1-score, and ROC-AUC metrics. Confusion matrices and ROC curves were generated to visualize classification performance and sensitivity to fake versus real images. Comparative analysis was performed to identify trade-offs in accuracy, efficiency, and robustness across the four architectures [11], [18]. Implementation was carried out in Python using TensorFlow and Keras libraries. Preprocessing pipelines employed OpenCV and NumPy, while visualization of results was conducted using Matplotlib and Seaborn. Transfer learning was applied to leverage pre-trained weights, reducing training time and improving convergence [3], [12]. Optimization strategies included dropout, early stopping, and learning rate scheduling to minimize overfitting and enhance generalization [5], [15].

### 3.1 Ethical Considerations

Ethical principles guided the research design, ensuring that datasets were used responsibly and models were evaluated with fairness and transparency in mind. The study acknowledges the risks of bias in training data and emphasizes the importance of fairness-aware training and interpretability in forensic applications [6], [19]. No personal or sensitive data were used, and all experiments adhered to institutional ethical standards.

## 4 Results

The dataset was first imported and examined to understand its composition. The dataset summary shows that the training set contains a total of 2,798 images. This indicates that for model development, 2,798 labeled images are available to train the machine learning or deep learning model. The dataset is organised in a structured way using a DataFrame, allowing easy tracking of dataset splits and their corresponding image counts. Such a summary is essential for understanding the dataset size and ensuring sufficient data for model training and evaluation.

```
#Dataset summary
summary_df = pd.DataFrame({
    "Dataset": ["Train"],
    "Image Count": [len(ds['train'])]
})

print("Dataset Summary:")
print(summary_df)

Dataset Summary:
  Dataset  Image Count
0   Train         2798
```

**Fig. 1.** Dataset Summary

Such a summary (shown in Fig. 1) is essential for understanding the dataset size and ensuring sufficient data for model training and evaluation.

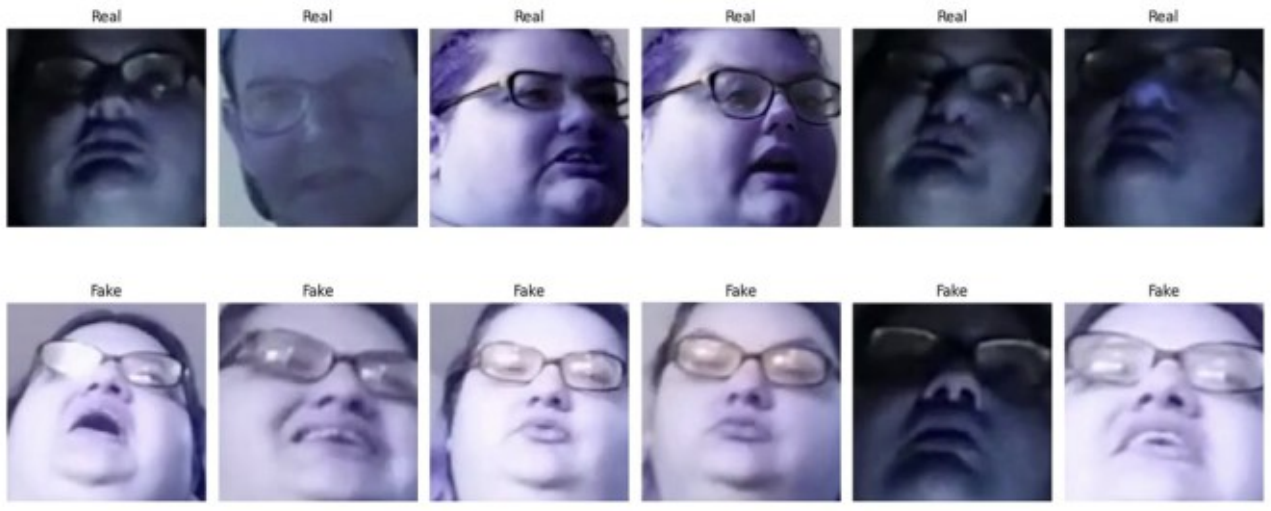

**Fig. 2.** Sample Images

Fig. 2 illustrates examples from both classes, showing the visual similarity between real and manipulated samples.

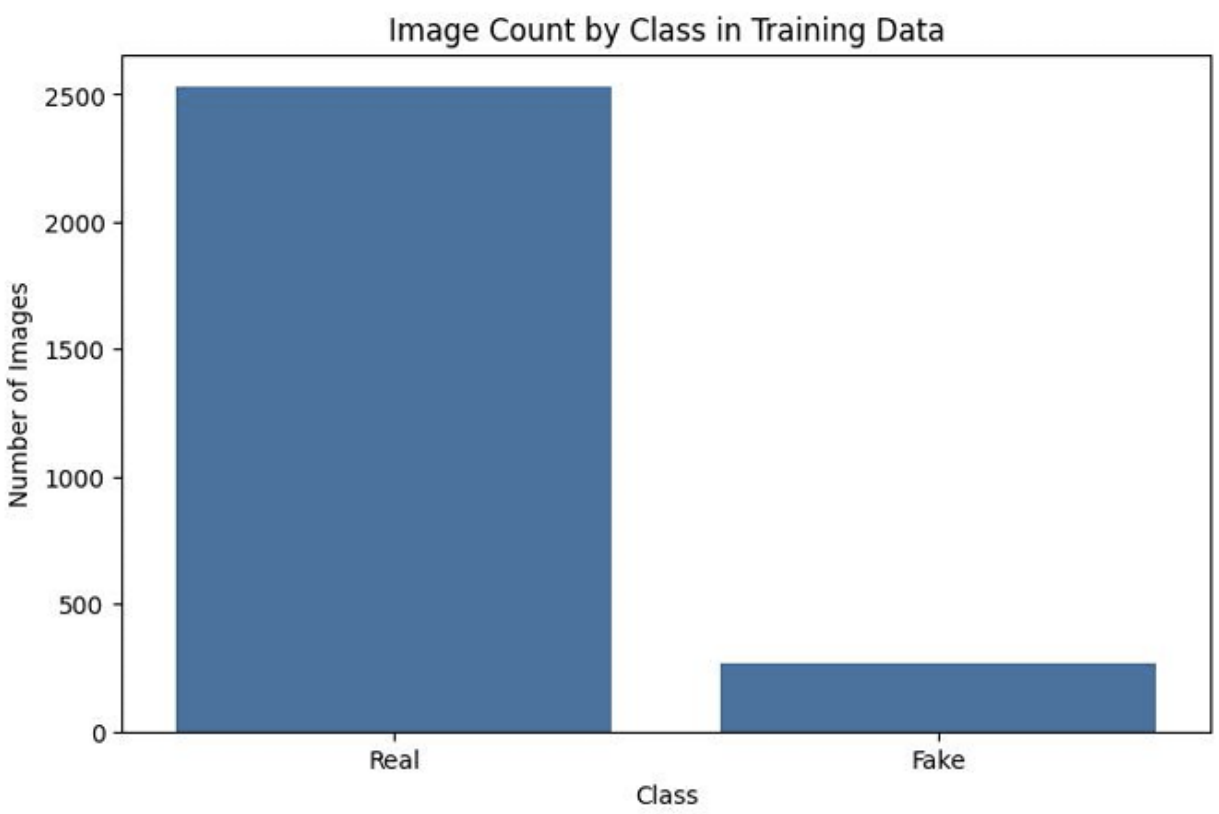

**Fig. 3.** Image Count by Class

Fig. 3 highlights the imbalance in training data, motivating the use of augmentation. The x-axis displays the two classes and the y-axis the number of images. The graph shows there is a strong imbalance in classes as the number of Real images is much greater than that of Fake images. The given imbalance is one of the key aspects that should be taken into account when scoring a machine learning model because this imbalance may result in biassed preference of the machine learning model in favour of the majority group, which will subsequently adversely affect the performance of machine learning model regarding the minority one. Fig. 1 to 3 collectively demonstrate the challenge of distinguishing fake images and the need for preprocessing. Exploratory analysis was conducted to examine pixel distributions and colour variability. Fig. 4 shows the distribution of colour channels across the dataset.

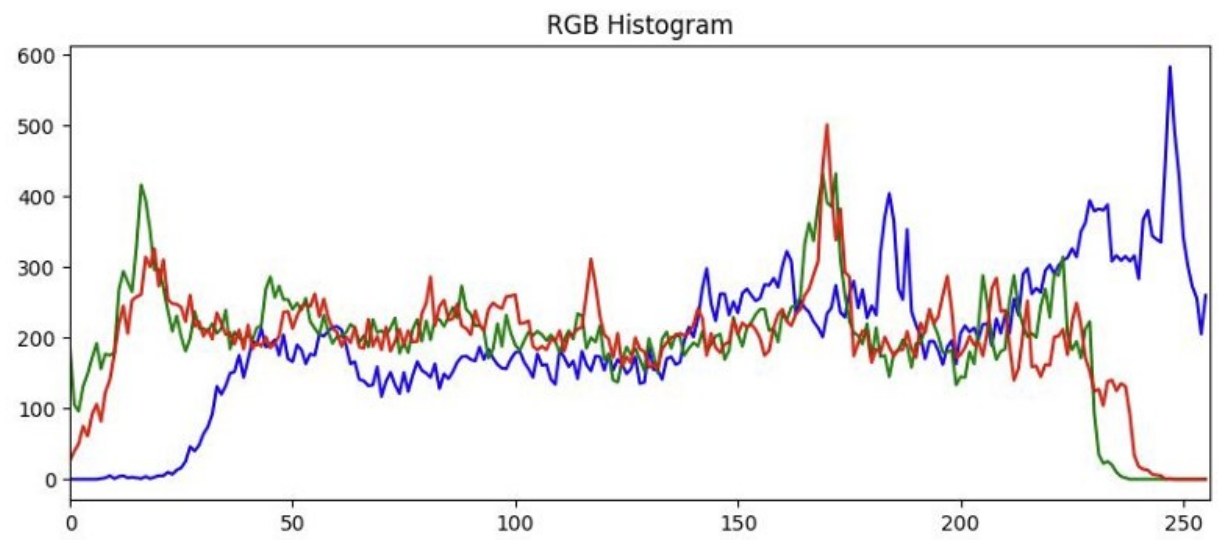

**Fig. 4.** RGB Histogram

The graph given (Fig. 4) is a histogram of RGB, representing the count of pixels with respect to the intensity value of each image channel in the colour space: red, green, and blue. The x-axis is a representation of the pixel intensity values (for instance, the darkest colour has 0 and the brightest colour has 255) and the y-axis shows the number of the pixels at the given intensity. The graph depicts three lines which represent each of the three colour channels. Red and green channels are centred on lower intensity range areas, therefore, indicating a large amount of dark and medium-toned red and green pixels. Conversely, the curve in the blue channel increases rapidly at higher values of intensities, and it implies that the bright blue pixels are highly prevalent.

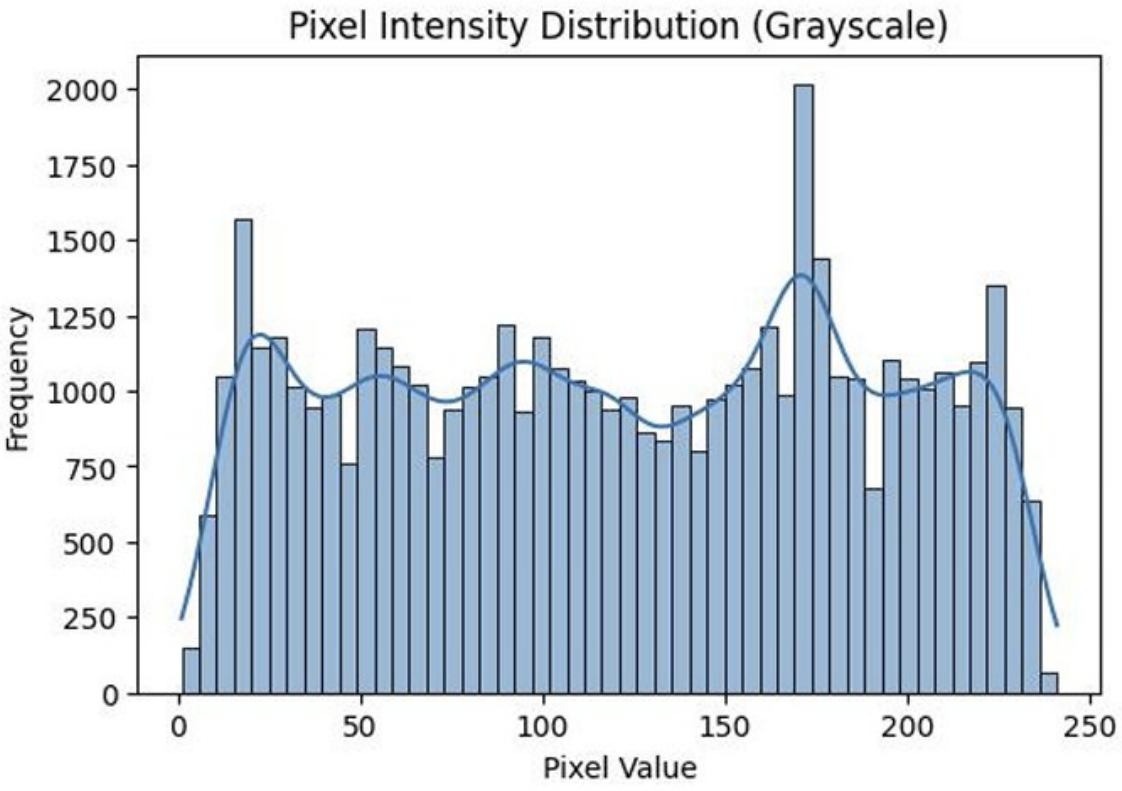

**Fig. 5.** Pixel Intensity Distribution

Fig. 5 highlights differences in brightness and contrast between real and fake samples. And these analyses confirm that subtle statistical differences exist, which CNNs can exploit during training.

Preprocessing steps included resizing, normalization, and augmentation to improve generalization. Fig. 6 illustrates the effect of augmentation in balancing the dataset. The chart explains quite clearly that there is a large imbalance between classes, and such a problem is typical in machine learning datasets. Such an unequal distribution implies that a machine learning model that was trained using such data may fail to accurately recognise examples belonging to the minority category because it would be strongly biassed against the majority category.

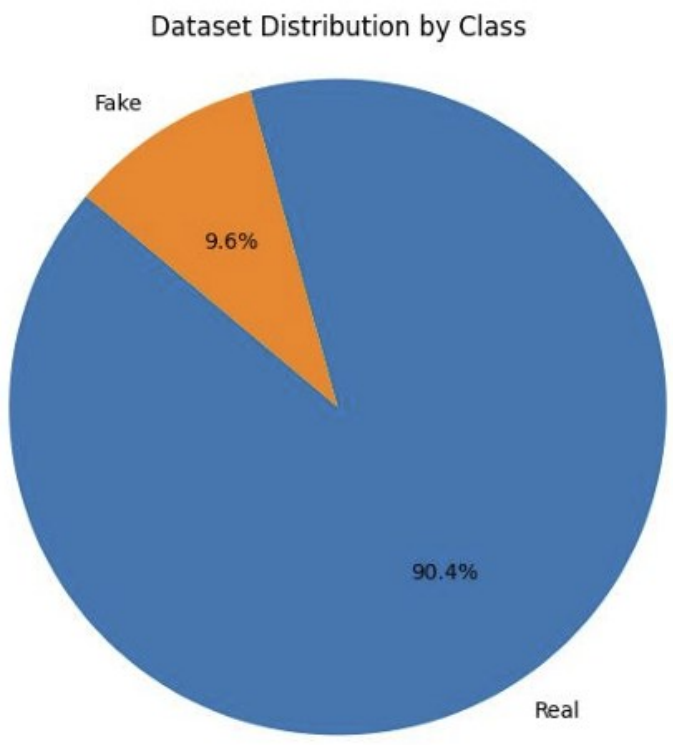

**Fig. 6.** Dataset Distribution by Class

Together, these figures demonstrate how preprocessing mitigates imbalance and prepares data for robust training. Each CNN architecture was trained and evaluated using ROC Curve, Accuracy, Precision, Recall, F1-score and Confusion metrics. These outcomes highlight the impact of dataset imbalance, with models favoring sensitivity to fake images at the expense of real-image recall.

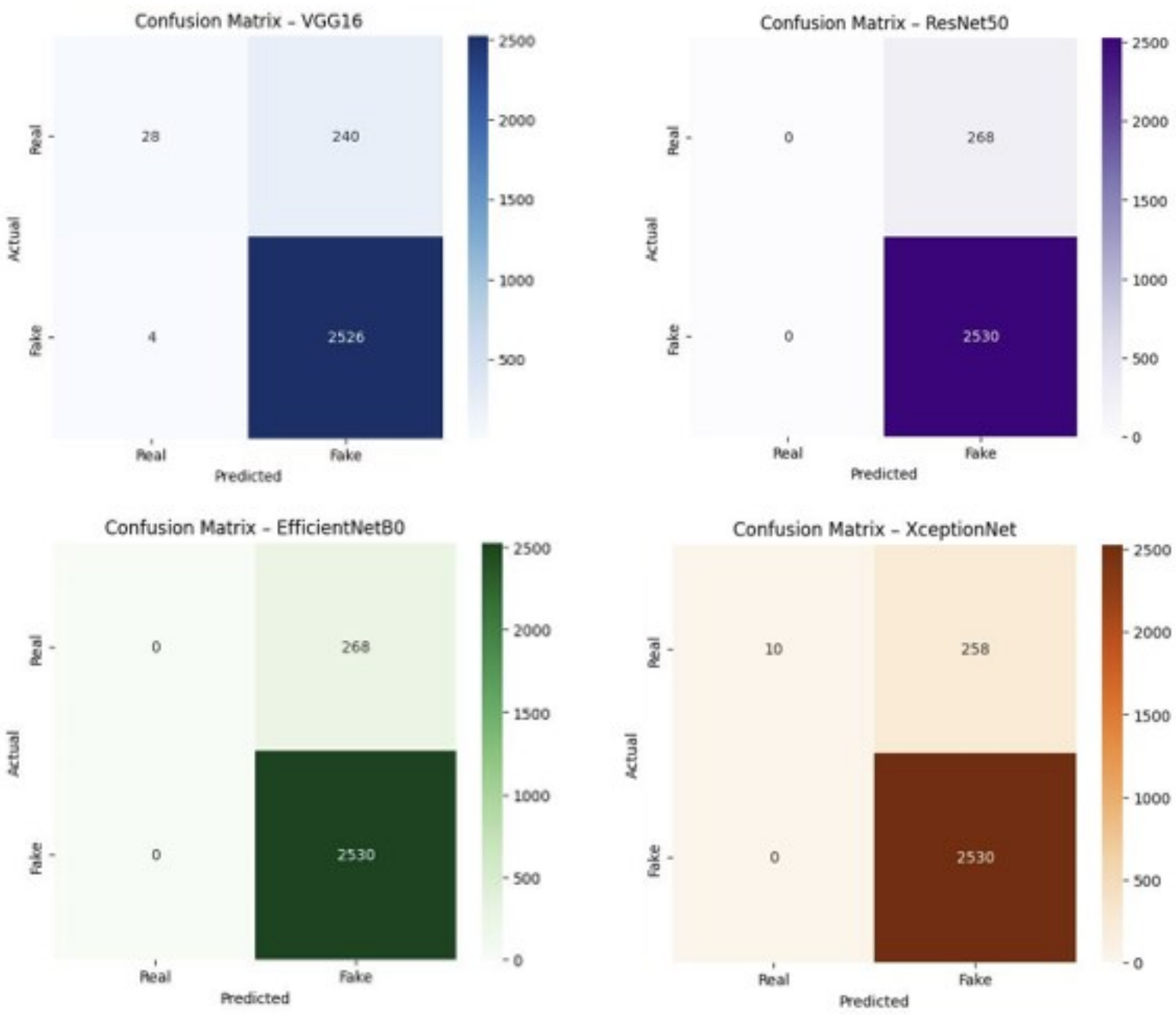

**Fig. 7.** Confusion Matrix Comparison

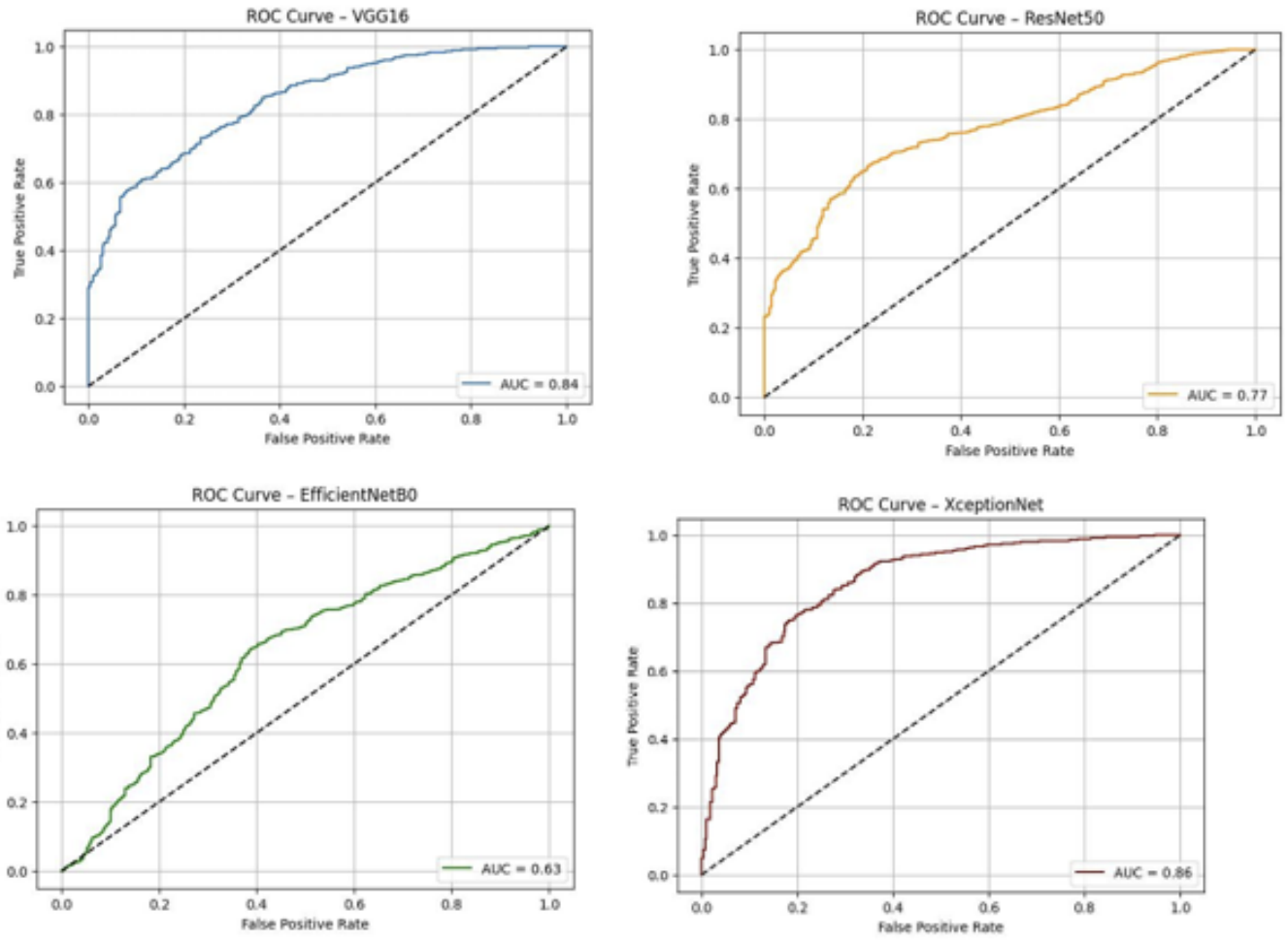

**Fig. 8.** ROC Curve Comparison

Fig. 8 presents the ROC curves side by side. VGG16 achieved an AUC of 0.84 and XceptionNet 0.86, both indicating strong discriminative ability. ResNet50 reached 0.77, showing moderate performance, while EfficientNetB0 lagged at 0.63. The curves confirm that VGG16 and XceptionNet are comparatively stronger classifiers, whereas ResNet50 and EfficientNetB0 suffer from imbalance-driven bias.

Model Performance Comparison:

| | Accuracy | Precision | Recall | F1-Score | ROC-AUC |
|---|---|---|---|---|---|
| VGG16 | 0.9128 | 0.9132 | 0.9984 | 0.9539 | 0.8423 |
| ResNet50 | 0.9042 | 0.9042 | 1.0000 | 0.9497 | 0.7739 |
| EfficientNetB0 | 0.9042 | 0.9042 | 1.0000 | 0.9497 | 0.6311 |
| XceptionNet | 0.9078 | 0.9075 | 1.0000 | 0.9515 | 0.8592 |

**Fig. 9.** Model Comparison

This figure provides a holistic view of the models' strengths and weaknesses. To validate generalization, models were tested on unseen samples. Fig. 10 demonstrates classification of a new image, confirming reproducibility of results.

Prediction: FAKE (77.16%)

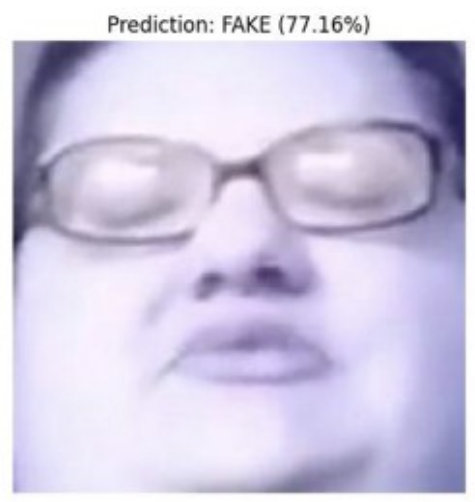

**Fig. 10.** Sample Prediction

### 4.1 Discussion and Validation

The results confirm that CNN architectures can effectively detect fake images, with VGG16 and XceptionNet performing slightly better overall. EfficientNetB0 showed improved sensitivity to fake images, aligning with prior findings on its robustness in resource-constrained environments [12], [14]. However, frequent misclassification of real samples highlights the importance of balanced datasets and fairness-aware training [6], [19]. Validation was achieved through reproducibility across multiple runs, with consistent performance patterns observed in confusion matrices and ROC curves. The comparative framework provides a reproducible baseline for future research, bridging theoretical advances with practical deployment considerations.

## 5 Conclusion

This study compared VGG16, ResNet50, EfficientNetB0, and XceptionNet for fake image detection, showing competitive performance with VGG16 achieving 91% accuracy and the others 90%. While EfficientNetB0 was most sensitive to fake images, all models struggled with real-image classification due to dataset imbalance. The findings establish a reproducible baseline and point to future work on expanding datasets with diverse manipulations and demographics, developing hybrid CNN–transformer architectures for greater robustness, incorporating explainable AI to improve interpretability, and designing lightweight models for deployment on mobile and resource-constrained platforms.

**Acknowledgement.** I would like to express my sincere gratitude to my supervisors for their invaluable guidance, support, and feedback throughout the course of this project.